\documentclass[letterpaper, 10pt, conference]{ieeeconf}      %

\IEEEoverridecommandlockouts                              %

\let\labelindent\relax                                    %

\usepackage{graphicx}            %
\usepackage{color}
\usepackage{amsmath}             %
\usepackage{amssymb}             %
\usepackage{amsfonts}            %
\usepackage{enumitem}            %
\usepackage{multirow}            %
\usepackage{siunitx}             %
\usepackage{url}                 %
\usepackage{xspace}              %
\usepackage[T1]{fontenc}         %
\usepackage[hidelinks]{hyperref} %
\usepackage{balance}
\usepackage[bottom]{footmisc} 
\usepackage{xfrac}
\usepackage{tensor}
\usepackage{todonotes}
\usepackage{tikz}
\usepackage{dblfloatfix}
\usetikzlibrary{matrix}

\usepackage[firstpage=true]{background}
\newcommand\copyrighttext{%
\parbox{\textwidth}{
\footnotesize
}}

\SetBgContents{\copyrighttext}
\SetBgScale{1}
\SetBgColor{black}
\SetBgAngle{0}
\SetBgOpacity{1}
\SetBgPosition{current page.north}
\SetBgVshift{-0.8cm}

\usepackage[bottom]{footmisc}

\graphicspath{{images/}}
\DeclareGraphicsExtensions{.pdf,.png,.jpg,.jpeg}

\newcommand{\mypm}{\mathbin{\mathpalette\@mypm\relax}}

\makeatletter
\newcommand{\amsray}{%
\mathpalette {\overarrow@\rayfill@}}
\def\rayfill@{\arrowfill@{\mkern4mu\mapstochar\relbar}\relbar{\mkern 4.08mu}}%
\makeatother

\newcommand{\seclabel}[1]{\label{sec:#1}}

\newcommand{\figlabel}[1]{\label{fig:#1}}
\newcommand{\tablabel}[1]{\label{tab:#1}}
\newcommand{\eqnlabel}[1]{\label{eqn:#1}}

\newcommand{\figref}[1]{Fig.~\ref{fig:#1}\xspace}
\newcommand{\tabref}[1]{Table~\ref{tab:#1}\xspace}
\newcommand{\eqnref}[1]{(\ref{eqn:#1})\xspace}

\title{\LARGE \textbf{Maximum Impulse Approach to Soccer Kicking for Humanoid Robots}}
\author{Grzegorz Ficht and Sven Behnke%
\thanks{All authors are with the Autonomous Intelligent Systems (AIS) Group, Computer Science Institute VI,
        University of Bonn, Germany. Email: {\tt\small ficht@ais.uni-bonn.de}.\newline
        This work has been supported by the German Federal Ministry of Education and Research (BMBF) grant 16ME0999 Robotics Institute Germany (RIG).}}

\usepackage{eso-pic}

\AtBeginDocument{\AddToShipoutPictureFG*{\AtTextUpperLeft{\put(0,\LenToUnit{9pt}){\parbox{\textwidth}{\centering\bfseries
Workshop on Humanoid Soccer Robots,\\ IEEE-RAS 23rd International Conference on Humanoid Robots (Humanoids), Nancy, France, November 2024.
}}}}}
\begin{document}

\bstctlcite{IEEEexample:BSTcontrol}

\maketitle
\thispagestyle{empty}
\pagestyle{empty}

\begin{abstract}

We introduce an analytic method for generating a parametric and constraint-aware kick for humanoid robots. 
The kick is split into four phases with trajectories stemming from equations of motion with constant acceleration.
To make the motion execution physically feasible, the kick duration alters the step frequency. 
The generated kicks seamlessly integrate within a ZMP-based gait, benefitting from the stability provided by the built-in controls.
The whole approach has been evaluated in simulation and on a real NimbRo-OP2X humanoid robot.

\end{abstract}
\section{Introduction}

Most approaches for in-walk kicking focus on generating smooth foot trajectories that overshoot the target step position mid-swing,
but are limited by not considering the dynamics and time necessary to perform the kick~\cite{yi2013improved,bormann2019developing,pavlichenko2022robocup,pavlichenko2023robocup}.
These issues can be accounted for when the kick is a self-contained motion sequence executed from a standing pose~\cite{behnke2008hierarchical}.
More recent work highlights the importance of building up momentum to deliver a powerful kick.
The approach presented by Marew et al.~\cite{marew2024biomechanics} builds a complex framework using retargeted motion-capture data, to 
generate optimal biomechanically-inspired kicks. As we show in this work, powerful kicks can be also obtained by using simple 
equations of motion that consider the physical constraints of the system.

\section{Approach}  \seclabel{approach}

Performing a powerful kick requires increasing the ball momentum $p$, defined as the impulse $J$. It can be maximized
by achieving maximum foot velocity at the moment of impact.
Assuming that the motion is mostly in the sagittal plane, this is synonymous with maximizing the hip and knee velocities. 
Aside from synchronizing their movement, the main challenge lies in reaching peak joint velocities
in the presence of physical constraints. These are mostly tied to the hip actuator as it needs to accelerate 
the whole mass of the leg for the kick. As the knee is placed lower in the kinematic chain, it actuates
only a portion of that mass with reduced leverage. Due to the relaxed knee joint constraints, we put focus 
on generating a physically feasible leg swing angle, which applies an offset to the hip motion. This generation is split into four subsequent phases: 
\textit{Prepare}, \textit{Swing}, \textit{Continue}, and \textit{Return}. As the kick trajectories 
apply offsets to the gait~\cite{ficht2023direct,ficht2023centroidal}, we derive them with respect to a zero
starting and ending angle $\theta_\textrm{l}$ and velocity $\omega_\textrm{l}$. %

\begin{figure}[t]
   \centering
   \centering
   \begin{tikzpicture}
       \matrix (m) [matrix of nodes, nodes={inner sep=-0.01cm, anchor=south west}, column sep=0, row sep=0] {
           \node {\includegraphics[height=2.5cm]{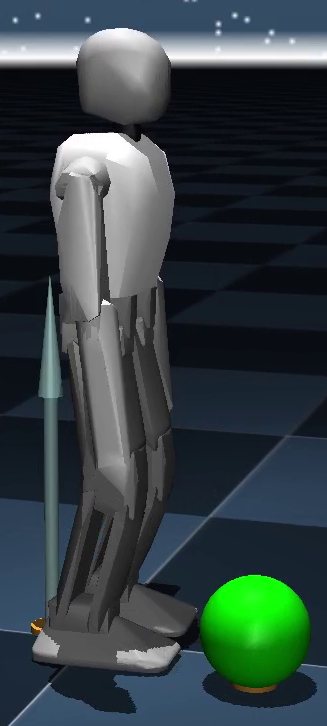}}; & 
           \node {\includegraphics[height=2.5cm]{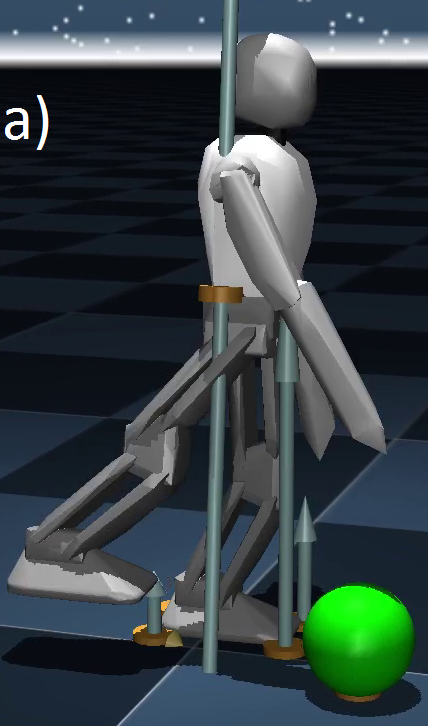}}; & 
           \node {\includegraphics[height=2.5cm]{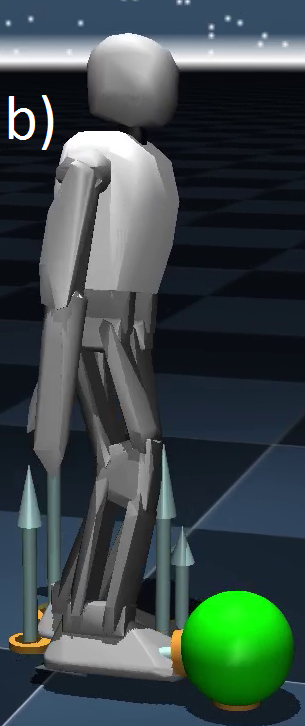}}; & 
           \node {\includegraphics[height=2.5cm]{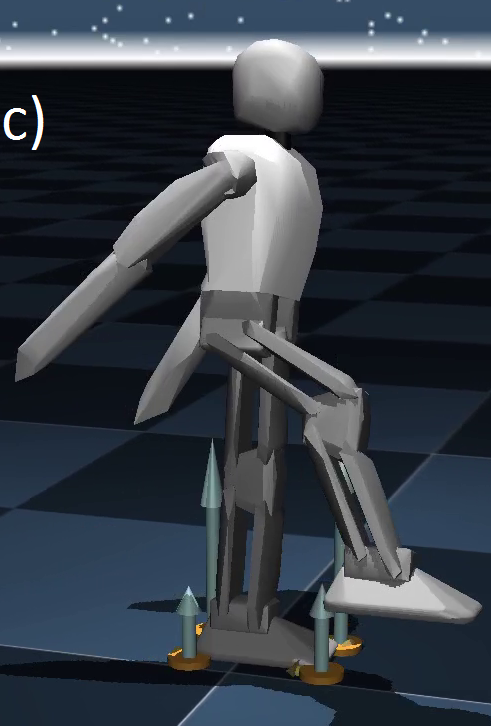}}; & 
           \node {\includegraphics[height=2.5cm]{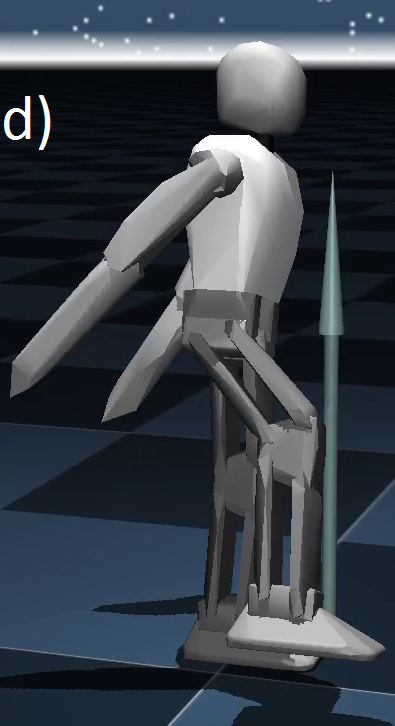}}; & 
           \node {\includegraphics[height=2.5cm]{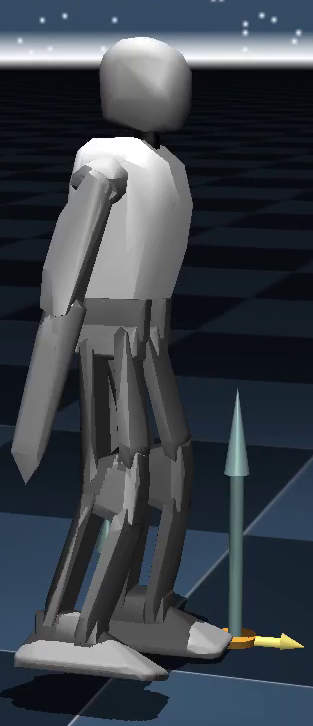}}; & 
           \node {\includegraphics[height=2.5cm]{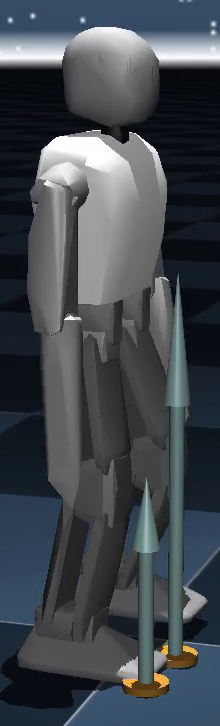}}; \\
       };
   \end{tikzpicture}\vspace{-2.3ex}
   \caption{Proposed maximum-impulse approach to kicking, consisting of four phases: a) preparation, b) swing, c) extension and d) return. 
 Throughout the kick, the robot remains in-gait.}
    \figlabel{kicking}\vspace{-2.8ex}
\end{figure}

\subsection{Swing Phase}

\begin{figure*}[!b]
	\vspace{-1ex}
   {\includegraphics[width=0.99\linewidth]{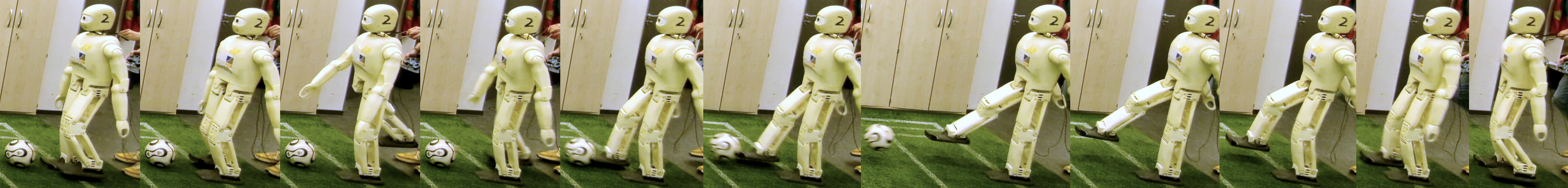}}
   \caption{NimbRo-OP2X kicking with the presented approach. From a walking state, the calculated swing motion builds up the leg velocity.
	At impact, the ball is visibly propelled into the air and travels the full \SI{5.5}{m} field in our lab, stopping at the goal. 
	After kicking, the robot continues walking despite the timing and impact disturbances to the gait.}
    \figlabel{kick_full}
	\vspace{-1ex}
\end{figure*}

Given a ball with radius $r_b$ and taking its distance to the tip of the foot $x_b$ %
and the average height of the hip origin $z_h$ throughout a gait cycle, we can compute the target kick 
angle $\theta_\textrm{k}$: 
\begin{equation} \eqnlabel{kickangle}
\theta_\textrm{k} = \text{atan2}(z_h-r_b, x_b-r_b).
\end{equation}
To reach the target kicking velocity $\omega_\textrm{k}$ from zero, the leg accelerates with 
$\alpha_\textrm{k}$ over the swing time $t_\textrm{sw}$:
\begin{equation} \eqnlabel{swingtime}
t_\textrm{sw} = \frac{\omega_\textrm{k}}{\alpha_\textrm{k}}.  
\end{equation}
This results in the leg traversing the angle $\theta_\textrm{sw}$, starting from the pre-swing $\theta_\textrm{pre}$:
\begin{equation} \eqnlabel{swingangle}
\theta_\textrm{sw} = \frac{1}{2}\alpha_\textrm{k}t_\textrm{sw}^2 , \qquad \theta_\textrm{pre} = \theta_\textrm{k} - \theta_\textrm{sw}.
\end{equation}
To maximize the impulse, $\omega_\textrm{k}$ needs to reach its peak value, which is synonymous 
with a maximum hip velocity $\omega_\textrm{h,max}$. Although it does not directly affect the momentum, 
it is also advantageous to minimize \eqnref{swingtime} by using maximum acceleration. 
This reduces the necessary single-support duration for kicking during the gait, 
at which the robot is most susceptible to losing balance. Another benefit is a smaller swing angle~$\theta_\textrm{sw}$, 
as larger ones might not always be feasible due to joint limits. Physically, our leg acceleration
is limited by the torque available at the hip $\tau_\textrm{h}$, and the leg inertia $I_\textrm{l}$:
\begin{equation} \eqnlabel{angacc}
\alpha_\textrm{k} = \frac{\tau_\textrm{h}}{I_\textrm{l}}. 
\end{equation}
The torque used for $\tau_\textrm{h}$ can be simply taken from the hip actuator specifications,
while $I_\textrm{l}$ needs to be computed from the desired leg configuration. In our framework,
we utilize a five-mass centroidal model~\cite{ficht2020fast}, from which we obtain $I_\textrm{l}$.

\subsection{Prepare Phase}

The preparation phase is only necessary if $\theta_\textrm{pre}$ is not \textit{on the way} to reach $\theta_\textrm{k}$,
in which case either the acceleration has to be reduced to match the target $t_\textrm{sw}$, or the leg can start 
accelerating to reach $\omega_\textrm{k}$ and continue until $\theta_\textrm{k}$ has been reached (with a new $t_\textrm{sw}$).
We will solely focus on the more demanding case with a necessary swing-up, where the leg first 
needs to reach a negative $\theta_\textrm{pre}$ and $\omega_{{\textrm{l}|\theta_\textrm{pre}}}=0$. We split this motion 
into symmetric acceleration and deceleration subphases, with a midpoint of $\theta_\textrm{pre}/2$.
The time necessary for the complete preparation phase is then:
\begin{equation} \eqnlabel{preptime}
t_\textrm{pre} = 2\sqrt{|\frac{\theta_\textrm{pre}}{\alpha_\textrm{k}}|}.
\end{equation}

	\begin{figure}[!t]
	\centering{\includegraphics[width=0.95\linewidth]{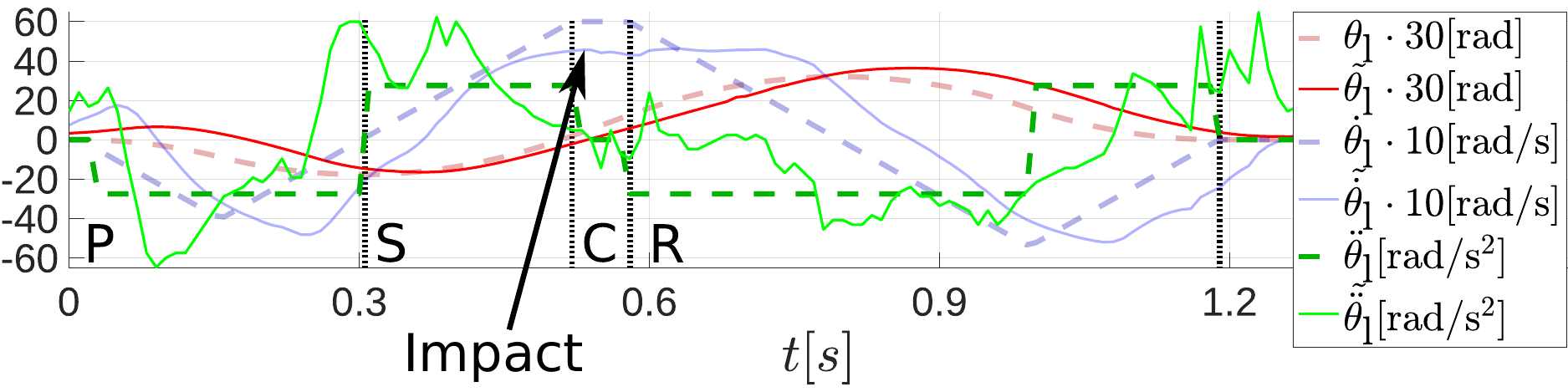}}
	\vspace{-2ex}
	\caption{Generated and measured\,(\~{}) leg motion trajectories during an experiment, with phases separated and denoted by their first letter.}
	\figlabel{waveforms} 
	\vspace{-4ex}
	\end{figure}

\subsection{Continue Phase}

The continuation phase serves to reduce sensitivity in the impact timing, stemming from inaccuracies in the 
control and ball position estimates. After reaching $\theta_\textrm{k}$ the leg continues moving with $\omega_\textrm{k}$
for a set extension angle of $\theta_\textrm{ext}$ or time $t_\textrm{ext}$ until it reaches $\theta_\textrm{ret}$:
\begin{equation} \eqnlabel{exttime}
\theta_\textrm{ret} = \theta_\textrm{k} + \theta_\textrm{ext}, \qquad t_\textrm{ext} = \frac{\theta_\textrm{ext}}{\omega_\textrm{k}}.
\end{equation}

\subsection{Return Phase}

After the kick, the leg needs to swiftly decelerate to return to the nominal gait trajectory. However, the
accumulated velocity will continue moving the leg forward despite decelerating. This happens until
the swing angle reaches its peak value $\theta_\textrm{post}$ with $\omega_{\textrm{l}|\theta_\textrm{post}}=0$. This
is an inverse of the swing phase trajectory, meaning that the velocity integral is the same:
\begin{equation} \eqnlabel{retangle}
\theta_\textrm{post} = \theta_\textrm{ret} + \theta_\textrm{sw}.
\end{equation}
After the leg stops at $\theta_\textrm{post}$, the return motion is split up into two subphases similarly to the 
preparation phase. The leg continues decelerating until it reaches $\theta_\textrm{post}/2$, after which it accelerates to 
return to the final zero angle with zero velocity. The time necessary for the complete return phase equals:
\begin{equation} \eqnlabel{rettime}
t_\textrm{ret} = t_\textrm{sw}+2\sqrt{|\frac{\theta_\textrm{post}}{\alpha_\textrm{k}}|}.
\end{equation}

\subsection{In-walk Application}

To execute the kick, the supporting phase of the gait needs to last the minimum required kicking time $t_\textrm{k}$. 
This is simply the sum of \eqnref{swingtime}, \eqnref{preptime}, \eqnref{exttime} and \eqnref{rettime}.
The gait frequency $f_\textrm{g}$ for the next step is set accordingly:
\begin{equation} \eqnlabel{stepfreq}
t_\textrm{k} = t_\textrm{pre} + t_\textrm{sw} + t_\textrm{ext} + t_\textrm{ret}, \qquad f_\textrm{g} = 1/t_\textrm{k}.
\end{equation}
Given the time elapsed since starting the kick and its phase parameters, we compute the swing angle $\theta_\textrm{l}$ 
using the constant acceleration-based equations of motion. The sagittal foot offsets $(x_\textrm{o},z_\textrm{o})$ are then computed as:
\begin{equation} \eqnlabel{kickcoord}
x_\textrm{o} = (z_\textrm{h}-r_\textrm{b})\text{sin}(\theta_\textrm{l}),\quad z_\textrm{o} = (z_\textrm{h}-r_\textrm{b})(1-\text{cos}(\theta_\textrm{l})),
\end{equation}
with the option of adding modifiers to synchronize the knee motion through extension and contraction based on $\theta_\textrm{l}$.

\section{Experimental Results}

The proposed approach was quantitatively verified with an accurate MuJoCo simulation of a \SI{135}{cm} tall, NimbRo-OP2X humanoid robot~\cite{ficht2018nimbro},
including inertias, joint limits and parallel linkage connections. An instance of the kick is shown in~\figref{kicking}, with corresponding swing trajectories in~\figref{waveforms}. 
We perform 10 \textit{on the spot} in-walk kicks with the presented approach,
and compare it to the waveform-based approach~\cite{pavlichenko2023robocup} used up to now. In none of the trials did the robot fall, 
despite slowing down $f_\textrm{g}$ from \SI{2.4}{Hz} to \SI{0.7}{Hz}, which noticeably disturbs the rhythm of the gait. 
\tabref{kickdist} reports the results. On average, our maximum-impulse kick propels the ball \SI{42}{\%} further, validating the efficacy of the approach.
Our method was also evaluated on hardware, with NimbRo-OP2X kicking across the full \SI{5.5}{m} 
length of the soccer field in our lab (see \figref{kick_full}). The actual distance of the kick would have been greater, as the ball hit the goal with some momentum.
\begin{table}[!t]
\renewcommand{\arraystretch}{1.2}
\caption{Kicking distance comparison in simulation}
\tablabel{kickdist}
\centering
\footnotesize
\begin{tabular}{c c c c}
\hline

\hline
$\quad$Kicking approach $\quad$&$\quad$mean / SD (m)$\,$&$\,$min$\,$&$\,$ max\\
\hline
\hline
$\quad$ Waveform-based~\cite{pavlichenko2023robocup} $\quad$&$\quad$ 5.28 / 0.31$\,$ & 4.79 & 5.69\\
$\quad$ Ours $\quad$&$\quad$  \bf{7.53} / 0.42$\,$ & \bf{6.79} & \bf{7.98} \\
\hline
\end{tabular}
\vspace{-4ex}
\end{table}

\section{Conclusions}

We presented an impulse-maximizing approach to kicking, incorporating joint velocity and torque limits in the planning.
It benefits from integrated balance controllers, as it builds on top of a functioning ZMP-based gait using a centroidal model.
The approach is parameterizable, as desired kicking velocities and ball positions can be set. 
The derivations also hold for the lateral plane allowing to extend the kick to be omnidirectional, which
will be a further focus of our work.

\balance
\bibliographystyle{IEEEtran}
\bibliography{wskick}

% Generated by IEEEtran.bst, version: 1.14 (2015/08/26)
\begin{thebibliography}{10}
\providecommand{\url}[1]{#1}
\csname url@samestyle\endcsname
\providecommand{\newblock}{\relax}
\providecommand{\bibinfo}[2]{#2}
\providecommand{\BIBentrySTDinterwordspacing}{\spaceskip=0pt\relax}
\providecommand{\BIBentryALTinterwordstretchfactor}{4}
\providecommand{\BIBentryALTinterwordspacing}{\spaceskip=\fontdimen2\font plus
\BIBentryALTinterwordstretchfactor\fontdimen3\font minus
  \fontdimen4\font\relax}
\providecommand{\BIBforeignlanguage}[2]{{%
\expandafter\ifx\csname l@#1\endcsname\relax
\typeout{** WARNING: IEEEtran.bst: No hyphenation pattern has been}%
\typeout{** loaded for the language `#1'. Using the pattern for}%
\typeout{** the default language instead.}%
\else
\language=\csname l@#1\endcsname
\fi
#2}}
\providecommand{\BIBdecl}{\relax}
\BIBdecl

\bibitem{yi2013improved}
S.-J. Yi, S.~McGill, and D.~D. Lee, ``Improved online kick generation method
  for humanoid soccer robots,'' in \emph{The 8th Workshop on Humanoid Soccer
  Robots, IEEE-RAS International Conference on Humanoid Robots (Humanoids)},
  2013.

\bibitem{bormann2019developing}
F.~Bormann, T.~Engelke, and F.-T. Sell, ``Developing a reactive and dynamic
  kicking engine for humanoid robots,'' Technical report, Universit{\"a}t
  Hamburg, Tech. Rep., 2019.

\bibitem{pavlichenko2022robocup}
D.~Pavlichenko, G.~Ficht, A.~Amini, M.~Hosseini, R.~Memmesheimer,
  A.~Villar-Corrales, S.~M. Schulz, M.~Missura, M.~Bennewitz, and S.~Behnke,
  ``{RoboCup 2022 AdultSize winner NimbRo: Upgraded Perception, Capture Steps
  Gait and Phase-Based In-Walk Kicks},'' in \emph{Robot World Cup XXV}.\hskip
  1em plus 0.5em minus 0.4em\relax {LNCS (LNAI), vol. 13561}, Springer, 2022,
  pp. 240--252.

\bibitem{pavlichenko2023robocup}
D.~Pavlichenko, G.~Ficht, A.~Villar-Corrales, L.~Denninger, J.~Brocker,
  T.~Sinen, M.~Schreiber, and S.~Behnke, ``{RoboCup 2023 Humanoid AdultSize
  Winner NimbRo: NimbRoNet3 Visual Perception and Responsive Gait with Waveform
  In-Walk Kicks},'' in \emph{Robot World Cup XXVI}.\hskip 1em plus 0.5em minus
  0.4em\relax {LNCS (LNAI), vol. 14140}, Springer, 2023, pp. 337--349.

\bibitem{behnke2008hierarchical}
S.~Behnke and J.~St{\"u}ckler, ``Hierarchical reactive control for humanoid
  soccer robots,'' \emph{International Journal of Humanoid Robotics (IJHR)},
  vol.~5, no.~03, pp. 375--396, 2008.

\bibitem{marew2024biomechanics}
D.~Marew, N.~Perera, S.~Yu, S.~Roelker, and D.~Kim, ``A biomechanics-inspired
  approach to soccer kicking for humanoid robots,'' \emph{arXiv preprint
  arXiv:2407.14612}, 2024.

\bibitem{ficht2023direct}
G.~Ficht and S.~Behnke, ``{D}irect {C}entroidal {C}ontrol for {B}alanced
  {H}umanoid {L}ocomotion,'' in \emph{Climbing and Walking Robots Conference
  (CLAWAR)}.\hskip 1em plus 0.5em minus 0.4em\relax Springer, 2023, pp.
  242--255.

\bibitem{ficht2023centroidal}
G.~Ficht and S.~Behnke, ``Centroidal state estimation and control for
  hardware-constrained humanoid robots,'' in \emph{IEEE-RAS International
  Conference on Humanoid Robots (Humanoids)}, 2023.

\bibitem{ficht2020fast}
G.~Ficht and S.~Behnke, ``Fast whole-body motion control of humanoid robots
  with inertia constraints,'' in \emph{IEEE International Conference on
  Robotics and Automation (ICRA)}, 2020, pp. 6597--6603.

\bibitem{ficht2018nimbro}
G.~Ficht, H.~Farazi, A.~Brandenburger, D.~Rodriguez, D.~Pavlichenko,
  P.~Allgeuer, M.~Hosseini, and S.~Behnke, ``Nimb{R}o-{OP2X}: Adult-sized
  open-source {3D} printed humanoid robot,'' in \emph{IEEE-RAS International
  Conference on Humanoid Robots (Humanoids)}, 2018.

\end{thebibliography}

\end{document}